\documentclass[journal,ITS]{IEEEtran}
\IEEEoverridecommandlockouts

\ifCLASSOPTIONcompsoc
  \usepackage[nocompress]{cite}
\else
  \usepackage{cite}
\fi

\ifCLASSINFOpdf
\else
\fi

\usepackage{cite}
\usepackage{amsmath,amssymb,amsfonts}
\usepackage{algorithmic}
\usepackage{array}
\usepackage{graphicx}
\usepackage{textcomp}
\usepackage[table]{xcolor}
\usepackage{multirow}
\usepackage{caption}
\usepackage{subcaption}

\def\BibTeX{{\rm B\kern-.05em{\sc i\kern-.025em b}\kern-.08em
    T\kern-.1667em\lower.7ex\hbox{E}\kern-.125emX}}
\usepackage{listings}

\definecolor{codegreen}{rgb}{0,0.6,0}
\definecolor{codegray}{rgb}{0.5,0.5,0.5}
\definecolor{codepurple}{rgb}{0.58,0,0.82}
\definecolor{backcolour}{rgb}{0.95,0.95,0.92}

\lstdefinestyle{mystyle}{
    backgroundcolor=\color{backcolour},   
    commentstyle=\color{codegreen},
    keywordstyle = {\color{magenta}},
    keywordstyle = [2]{\color{lime}},
    keywordstyle = [3]{\color{yellow}},
    keywordstyle = [4]{\color{teal}},
    numberstyle=\tiny\color{codegray},
    stringstyle=\color{codepurple},
    basicstyle=\ttfamily\footnotesize,
    breakatwhitespace=false,         
    breaklines=true,                 
    captionpos=b,                    
    keepspaces=true,                 
    numbers=left,                    
    numbersep=5pt,                  
    showspaces=false,                
    showstringspaces=false,
    showtabs=false,                  
    tabsize=2
}

\title{Fast-COS: A Fast One-Stage Object Detector Based on Reparameterized Attention Vision Transformer for Autonomous Driving \\
}
\author{Novendra Setyawan, Ghufron Wahyu Kurniawan, Chi-Chia Sun, \IEEEmembership{Member, IEEE}, \\
Wen-Kai Kuo, \IEEEmembership{Member, IEEE}, Jun-Wei Hsieh \IEEEmembership{Senior Member, IEEE} \\
\IEEEauthorblockA{} 
\thanks{Novendra Setyawan and Wen-Kai Kuo are with Department of Electro-Optics, National Formosa University, Taiwan;

Novendra Setyawan also with Department of Electrical Engineering University of Muhammadiyah Malang, Indonesia; 

Ghufron Wahyu Kurniawan is with Department of Electrical Engineering, National Formosa University, Taiwan; 

Chi-Chia Sun is with Department of Electrical Engineering, National Taipei University, Taiwan; 

Jun-Wei Hsieh is with College of Artificial Intelligence and Green Energy, National Yang Ming Chiao Tung University, Taiwan; 

Corresponding Author is Chi-Chia Sun (\textit{E-mail: chichiasun@gm.ntpu.edu.tw})
}
}

\begin{document}
\maketitle
\markboth{Under Review on IEEE Transactions}%
{Setyawan \etal{}: Fast-COS}

\begin{abstract}
The perception system is a core element of an autonomous driving system, playing a critical role in ensuring safety. The driving scene perception system fundamentally represents an object detection task that requires achieving a balance between accuracy and processing speed. Many contemporary methods focus on improving detection accuracy but often overlook the importance of real-time detection capabilities when computational resources are limited. Thus, it is vital to investigate efficient object detection strategies for driving scenes. This paper introduces Fast-COS, a novel single-stage object detection framework crafted specifically for driving scene applications. The research initiates with an analysis of the backbone, considering both macro and micro architectural designs, yielding the Reparameterized Attention Vision Transformer (RAViT). RAViT utilizes Reparameterized Multi-Scale Depth-Wise Convolution (RepMSDW) and Reparameterized Self-Attention (RepSA) to enhance computational efficiency and feature extraction. In extensive tests across GPU, edge, and mobile platforms, RAViT achieves 81.4\% Top-1 accuracy on the ImageNet-1K dataset, demonstrating significant throughput improvements over comparable backbone models such as ResNet, FastViT, RepViT, and EfficientFormer. Additionally, integrating RepMSDW into a feature pyramid network forms RepFPN, enabling fast and multi-scale feature fusion. Fast-COS enhances object detection in driving scenes, attaining an AP50 score of 57.2\% on the BDD100K dataset and 80.0\% on the TJU-DHD Traffic dataset. It surpasses leading models in efficiency, delivering up to 75.9\% faster GPU inference and 1.38× higher throughput on edge devices compared to FCOS, YOLOF, and RetinaNet. These findings establish Fast-COS as a highly scalable and reliable solution suitable for real-time applications, especially in resource-limited environments like autonomous driving systems.
\end{abstract}
\section{Introduction}
\label{sec:intro}
Autonomous driving systems represent a major breakthrough in transportation by allowing vehicles to operate without human involvement. These systems usually integrate various sensors as perception systems to collect real-time traffic data, enabling independent navigation \cite{betz2022autonomous}. Cameras supply essential high-resolution visual data needed for driving scene image processing tasks like object detection \cite{wang2024occludedinst}. Robust perception of the driving environment is crucial for autonomous vehicles in complex scenarios. This requires high accuracy, real-time processing, and resilience. It is crucial to accurately recognize and predict the movements of objects, performing efficient real-time processing to avoid decision-making delays that might lead to traffic congestion or accidents, and to ensure operation in poor weather or low-light conditions. An effective object detection algorithm is vital to the safety and effectiveness of camera-based perception systems in autonomous vehicles \cite{chib2023recent}.

In contemporary object detection architectures within deep learning, the structure typically comprises backbone, neck, and head detector elements \cite{chen2022mixed, wang2025yolov9, farhadi2018yolov3}. Detectors are mainly divided into two categories: two-stage and one-stage. Two-stage models, such as those in the R-CNN family \cite{ yang9847011, li10474576}, emphasize region proposal and feature extraction. These models are known for their precise object localization but face increased computational costs due to the numerous region proposals. Conversely, single-stage detectors like You Only Look Once (YOLO) \cite{Tian9298480} or RetinaNet \cite{ross2017focal} conduct object detection and localization regression in a single network pass. A variant of the single-stage detector is the Fully Convolutional One-Stage detector (FCOS) \cite{9010746,tian2020fcos}, which employs an anchor-free strategy by predicting on a per-pixel basis, thus removing the need for predefined anchor boxes and enhancing computational efficiency. Nevertheless, FCOS is criticized for having an inefficient model architecture due to a substantial backbone and neck. 

Throughout the decade, Convolutional Neural Networks (CNNs), notably ResNet \cite{he2016deep}, have been frequently employed as backbone networks due to their outstanding performance in numerous downstream tasks \cite{Djenouri9757754}, including biometric recognition \cite{hsu2024inpainting, li2021identity}, medical segmentation \cite{tan2023multi}, and image dehazing \cite{li2024ma}. Nevertheless, owing to the constraints of the receptive field and short-range dependencies, they encounter issues with occlusion, especially prevalent in driving scene object detection \cite{wang2023centernet}. Recently, vision models based on Transformers have demonstrated exceptional success as backbone networks \cite{liu2021swin} in various computer vision applications, or as segmentation encoder-decoder architectures \cite{chen2023transattunet}, leveraging their global receptive field and long-range dependencies, surpassing CNN performance. However, these models are generally computationally intensive due to their quadratic computational complexity. For instance, the original Vision Transformer (ViT) \cite{dosovitskiy2020image} necessitates 85 million to 632 million parameters for ImageNet classification. This complexity poses challenges for deployment on resource-constrained devices, such as mobile and edge devices, and may not be suitable for some applications, such as driving scene object detection \cite{wang2023centernet} and resource-limited platform deployment. 

Several efficient design approaches have been developed to make transformers efficient or can be implemented in mobile or edge devices \cite{maaz2022edgenext, li2022efficientformer, mehta2022mobilevit, mehta2023separable, li2023rethinking}. The most groundbreaking methods integrate transformers with convolutional neural networks (CNNs) \cite{maaz2022edgenext, mehta2022mobilevit}. Some strategies introduce a novel self-attention model with linear complexity \cite{mehta2023separable} and a dimension-sensitive architecture \cite{li2022efficientformer, li2023rethinking}. These methods demonstrate that CNNs play vital roles in deploying transformers on resource-constrained devices. Alternatively, there's a focus on designing a transformer-oriented vision model at the architectural level for rapid inference \cite{graham2021levit, liu2023efficientvit, yun2024shvit}. Many achieve expedited inference with very low resolution from the initial architectural phase using a 16 $\times$ 16 stem. Moreover, innovative self-attention mechanisms have been introduced to minimize computational redundancy \cite{liu2023efficientvit, yun2024shvit}. While these methods perform well in fast GPU inference, they are less effective for inference on resource-limited hardware with fewer core processors. 

In contrast, the sophisticated architecture of ViTs has inspired recent CNN development \cite{liu2022convnet, vasu2023fastvit, wang2024repvit}. Instead of using combined spatial and channel feature extraction, such as traditional CNN \cite{howard2019searching, sandler2018mobilenetv2}, they design it separately following ViT's architecture. To capture global spatial context, several CNNs try to increase the kernel size to $7 \times 7$ \cite{liu2022convnet, vasu2023fastvit}  instead of using the common $3 \times 3$ \cite{wang2024repvit}. Furthermore, in \cite{ding2022scaling, liu2022more}, they scaled the kernel to $51 \times 51$ to attain a larger receptive field. However, extremely large kernels significantly increase memory access and parameters, making optimization challenging.

To address the trade-off between performance and speed, we introduce RAViT, which denotes Reparameterized Attention Vision Transformer. This hybrid transformer incorporates a reconfigurable token mixer blending attention with multiscale large kernel CNN. On a macro scale, RAViT merges the general framework of lightweight ViT \cite{graham2021levit, liu2023efficientvit, yun2024shvit} with the latest CNN-based design \cite{wang2024repvit}. At the micro level, the Re-parameterized Multi Scale Depth Wise Convolution (RepMSDW) and Re-parameterized Self-Attention (RepSA) are introduced to maintain global and local dependencies. Utilizing the RAViT as the backbone, we optimize FCOS to be more efficient and fast, hence we call it Fast Convolutional One-Stage object detector or Fast-COS. Not only the backbone, in the neck level, the Feature Pyramid Network (FPN) of FCOS is enhanced with RepFPN, which takes advantage of RepMSDW. Extensive evaluations confirm its efficiency across various vision benchmarks, including ImageNet-1K for backbone image classification, and BDD100K and TJU-DHD Traffic for object detection in driving scenes. In summary, our key contributions include:
\begin{itemize}
    \item We propose RAViT, a hybrid vision transformer with multiscale large kernel reparameterization, integrating partial self-attention.
    \item We demonstrate the effectiveness of RAViT as a backbone network for extracting multi-scale features and highlight its potential to enhance the FCOS object detector, achieving high accuracy on the BDD100K and TJU-DHD datasets.
    \item By leveraging RepMSDW, we refined the original FCOS FPN neck for multi-scale feature extraction, showcasing RepFPN's role as the neck along with the RAViT backbone, forming the Fast-FCOS object detector to achieve superior accuracy in the driving scene datasets.
    \item We shows that our models exhibit low latency on various platforms, including mobile, edge devices, and desktop GPUs, which will significantly benefit driving scene object detection systems on diverse hardware.
\end{itemize}


\begin{figure*}
    \centering
    \includegraphics[width=17.6cm]{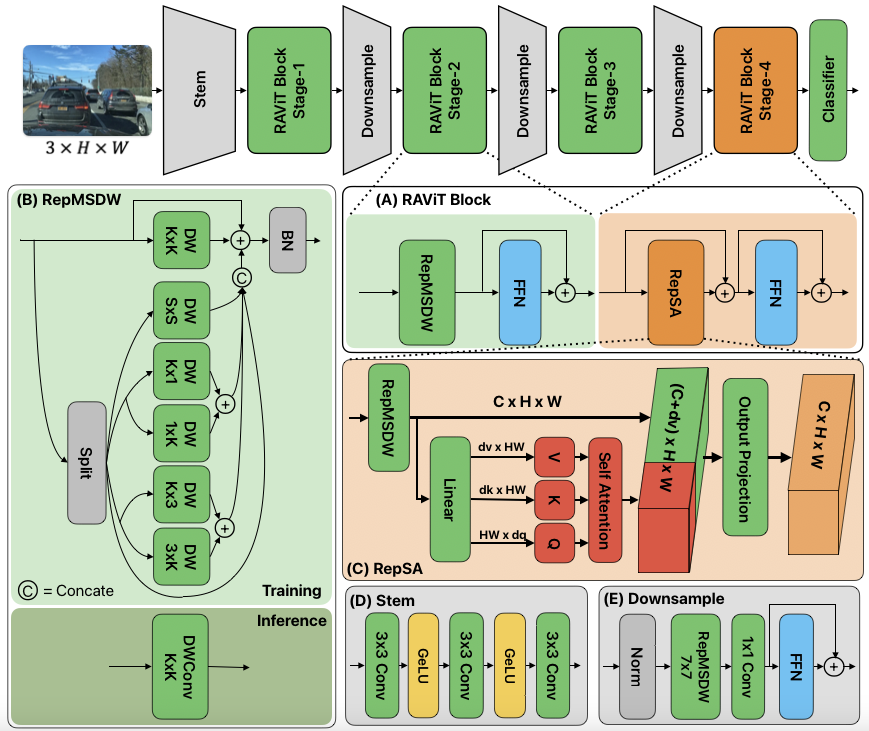}
    \caption{Overall RAViT backbone with 4-stage pyramid architecture with structural reparameterization to increase the inference speed. Stages 1 and 2 use the same RepMSDW as the spatial token mixer. (a) RAViT Block that can mix between RepMSDW and RepSA (b) Overview the reparameterization of multi-scale depth-wise convolution (RepMSDW) (c) Illustration of Reparameterize Self-Attention (d) Architecture of Stem (e) Architecture of Downsample }
    \label{fig:RAViT}
\end{figure*} 
\section{Related Works}
\subsection{Efficient Vision Transformer}
Recent advances in efficient vision transformers began with MobileViTs \cite{mehta2022mobilevit}, which combined MobileNets' efficiency with Vision Transformers' (ViTs) global modeling abilities. EfficientFormers \cite{li2022efficientformer, li2023rethinking} demonstrated a feature dimension-aware design that employs hardware-optimized 4D modules and effective 3D Multi-Head Self-Attention (MHSA) blocks. FastViT \cite{vasu2023fastvit} enhanced model capacity and efficiency by integrating 7 x 7 depthwise convolutions with Structural Re-Parameterization. EdgeNext \cite{maaz2022edgenext}applied local-global blocks to better combine MHSA and convolution. SHViT \cite{yun2024shvit} addressed computational redundancy through a Single-Head Self-Attention (SHSA) mechanism applied to a subset of channels. EMO \cite{zhang2023rethinking} addressed a simplicity that merged window self-attention with inverted bottleneck convolution into a single block.

\subsection{Large Kernel Convolution}
Initially, traditional CNNs like AlexNet and GoogLeNet favored large kernels in their early architecture, but with VGG \cite{ding2021repvgg}, the focus shifted to using stacked 3 × 3 kernels. InceptionNets \cite{szegedy2016rethinking, szegedy2017inception} improved computational efficiency by breaking down n×n convolutions into sequential $1 \times n$ and $n \times 1$ operations. SegNeXt \cite{guo2022segnext} extended the effective kernel size with combined $1 \times k + k \times 1$ and $k \times 1 + 1 \times k$ convolutions for semantic segmentation tasks. MogaNet \cite{li2023moganet} employed multi-scale spatial aggregation blocks that utilize dilated convolutions to capture discriminative features. ConvNeXt \cite{liu2022convnet} experimented with modern CNN designs using 7 × 7 depthwise convolutions, reflecting the Swin Transformer \cite{liu2021swin} architectural strategies. InceptionNeXt \cite{yu2024inceptionnext} enhanced throughput and performance by splitting large-kernel depthwise convolutions into four parallel branches. SKNet \cite{cui2021sknet} and LSKNet \cite{li2024lsknet} employed multi-branch convolutions in both channel and spatial dimensions. Moreover, RepLKNet \cite{ding2022scaling} expanded kernel sizes to 31 $\times$ 31 using SRP, achieving performance comparable to Swin Transformers. 

\subsection{Structural Re-Parameterization}
Recent studies, RepVGG \cite{ding2021repvgg} have shown that reparameterizing skip connections can lessen memory access costs. To improve efficiency, previous works such as MobileOne \cite{vasu2023mobileone} have utilized factorized k×k convolutions combined with depthwise or grouped convolutions followed by 1×1 pointwise convolutions. This approach significantly boosts overall model efficiency, although the reduced number of parameters could lead to lower capacity. Recently, reparameterized MLP like the token mixer that is proposed in \cite{ding2022repmlpnet} called RepMLPNet. To the best of our knowledge, using structural reparameterization to remove skip connections with multiple-scale convolution has not been explored in hybrid transformer architectures before. Moreover, the combination of reparameterized convolution combined with the self-attention mechanism has also not been explored yet.

\section{Proposed Method}
In this section, the RAViT hybrid transformer backbone is proposed. Firstly, we analyze the architecture at the macro level for the limited-resource hardware purpose. Then we develop the architecture at the micro-level. In  micro level, instead of using Multi-Head Self-Attention (MHSA) as the feature or token mixer that is computationally expensive, especially in high resolution, we proposed Reparameterized Multi-Scale Depth-Wise Convolution (RepMSDW). We also proposed Reparameterized Self-Attention (RepSA), which amalgamates the RepMSDW with self-attention to balance the local and global spatial understanding in features. Then, the RAViT will be utilized to improve the FCOS to perform the downstream task in driving scene object detection.
\subsection{Analysis in Macro Design}
\label{subsec:abl-macro}
Most of the recent macro-design of the vision transformer is based on a feature pyramid architecture of MetaFormer \cite{yu2022metaformer, yu2023metaformer} that stacks two residual blocks as shown in Fig. \ref{fig:RAViT}. The architecture starts with a stem module that might consist of a single \cite{yu2022metaformer} or sequence of two \cite{maaz2022edgenext, li2022efficientformer, wang2024repvit} or three blocks \cite{yun2024shvit, liu2023efficientvit} $3 \times 3$ convolutions with a stride of 2. The architecture in macro incorporates a token mixer block for spatial feature extraction, followed by a channel mixer block. Each block contains a normalization layer and residual or skip connection to steady the loss and advance the training process. Suppose $X_i, X'_i, X''_i \in \mathbb{R}^{H_i\times W_i\times C_i}$ are the feature maps in stage i with $H_i \times W_i$ resolution and $C_i$ channel number; the details of the block are explained in Equation \ref{eq:meta}.
\begin{equation}
    \begin{split}
    X'_i &= X_i + Norm(TokenMixer(X_i)) \\
    X''_i &= X'_i + Norm(ChannelMixer(X'_i))
    \end{split}
\label{eq:meta}
\end{equation}
$TokenMixer(.)$ operators commonly configured as convolution mixer or self-attention \cite{yu2023metaformer}. $ChannelMixer(.)$ block contains the feed-forward network (FFN) that is conducted by two linearly fully connected layers and a single activation function that can be expressed in Equation \ref{eq:ffn},
\begin{equation}
    FFN(X'_i)=\sigma(X'_iW_e+b_e)W_r+b_r
    \label{eq:ffn}
\end{equation}
where $W_e\in\mathbb{R}^{(C_i)\times rC_i}$ and $W_r\in\mathbb{R}^{(rC_i)\times C_i}$ are the layer weights, $r$ is the expansion ratio with a default value of 3, $b_e\in\mathbb{R}^{rC_i}$ and $b_r\in\mathbb{R}^{C_i}$ are bias weights of the fully connected layer. Operation $\sigma$ is chosen using the activation function $GELU(.)$ since it balances accuracy and inference speed according to \cite{wang2024repvit}. 

To construct an efficient yet low-cost model for mobile and edge devices, we analyze the architecture in the macro-design. First, incorporating $3 \times 3$ dwconv as a token mixer, we compare the 3-stage architecture with the $16 \times 16$ stem used in \cite{graham2021levit, liu2023efficientvit, yun2024shvit} and the 4-stage architecture with the $4 \times 4$ stem commonly used in \cite{vasu2023fastvit, li2022efficientformer, maaz2022edgenext}. As shown in Table \ref{tab:ablation_macro}, the comparison between V1 and V2 shows that even the 3 stage with the $16 \times 16$ stem can increase 3 $\times$ the GPU throughput. However, it does not have a significant effect on inference latency on both edge and mobile devices.  So, in the RAViT, we decided to use the 4-stage with the stem $4 \times 4$.
\subsection{Reparameterize Multi-Scale Depth-Wise Convolution}
Reparameterize Multi-Scale Depth-Wise Convolution (RepMSDW), as depicted in Fig. \ref{fig:RAViT} (b), is inspired by \cite{szegedy2017inception, yu2024inceptionnext}, which has several branch-depth-wise convolutions with different kernel sizes to expand the effective receptive field and feature extraction. By incorporating this multi-scale strategy, our model aims to replicate the multiple range modeling capabilities while maintaining locality and efficiency. The formulation of RepMSDW is described in Equation \ref{eq:repmsdw},
\begin{equation}
    \begin{split}
    X_1, & X_2, X_3, X_4 = Chunk(X_i) \\
    X_1 &= DWC_{s\times s}(X_1) \\
    X_2 &= DWC_{1\times k}(X_2) + DWC_{k\times 1}(X_2) \\
    X_3 &= DWC_{3\times k}(X_3) + DWC_{k\times 3}(X_3) \\
    X'_i &= BN (DWC_{k\times k}(X_i) + Cat([X_1, X_2, X_3, X_4])) 
\end{split}
\label{eq:repmsdw}
\end{equation}
where $X_i\in\mathbb{R}^{H_i\times W_i\times C_i}$ denotes the feature map in the stage-i that will be split along the channel dimension into $X_1, X_2, X_3, X_4 \in \mathbb{R}^{H_i\times W_i\times \frac{C_i}{4}}$. Each feature will be fed into different branches of depth-wise convolution (DWC), where $k$ denotes the largest kernel size and $s$ denotes the square kernel size equal to $k/2 \in \mathbb{N}=\{1,2,3,..\} $. Then each output branch is concatenated ($Cat(.)$) along the channel. Following the re-parameterization procedure in \cite{ding2021repvgg}, the sum between the concatenated branch and main $DWC_{k\times k}(.)$ with BatchNorm layer ($BN(.)$) will be re-parameterized into a single depth-wise convolution, denoted as $RepMSDW(.)$.

Table \ref{tab:ablation_token_mixer} shows the effectiveness of proposed RepMSDW. Compared to a single branch with square $k \times k$ and without reparameterization, the reparameterized multiple branch kernel has better accuracy without sacrificing the inference speed. Next, we try to increase the RepMSDW kernel size from $3 \times 3$ into $7 \times 7$ in the last two stages. As shown in Table \ref{tab:ablation_macro}, the increment of kernel size from V2 to V3 can increase the accuracy to 79.1\% with only a 2.5\% and 1\% latency drop in mobile and edge devices. Using the RepMSDW as a token mixer, our design has similar accuracy with FastViT-T12 \cite{vasu2023fastvit}; however, our approach has 26\% and 10\% faster inference speed in mobile and edge devices.

\begin{figure*}
    \centering
    \includegraphics[width=17cm]{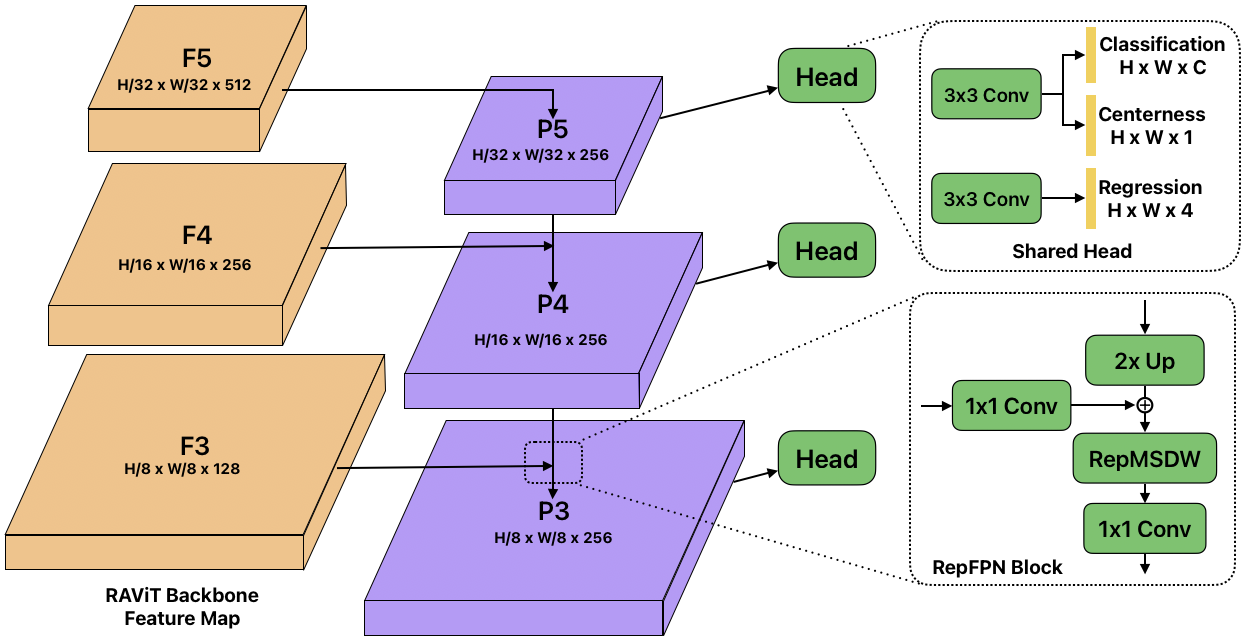}
    \caption{Fast-COS object detector utilize proposed RAViT backbone and RepFPN with Re-parameterized Multi-Scale DW Convolution and Re-parameterized Self Attention}
    \label{fig:RepFPN}
\end{figure*} 

\subsection{Reparameterize Self-Attention}
The reparameterized self-attention (RepSA) extends the spatial aggregation to reach long-range dependencies that become limitations of convolution. RepSA utilizes single-head self-attention with a quarter channel projected from RepMSDW, with details presented in Fig. \ref{fig:RAViT} (c). Generally, the formulation of RepSA is described in the following equation:
\begin{align}
    U &= RepMSDW(X_i) \\
    Q,\ K,\ V &=W_q*U,\ W_k*U,\ W_v*U\\
    Attn &= Softmax\Big(Q^TK/\sqrt{d_{q}}\Big)V \\
    RepSA(X_i) &=W_o*Cat(U,Attn)
\end{align}
where $U\in\mathbb{R}^{H_i\times W_i\times C_i}$, $Q\in\mathbb{R}^{d_q\times H_iW_i}$, $K\in\mathbb{R}^{d_k\times H_iW_i}$, and $V\in\mathbb{R}^{d_v\times H_iW_i}$. $W_q\in\mathbb{R}^{C_i\times d_q}$, $W_k\in\mathbb{R}^{C_i\times d_k}$, and $W_v\in\mathbb{R}^{C_i\times d_v}$ are denoted as the linear operation weights to project the feature $U$ into the query ($Q$), key ($K$), and value ($V$). The attention score $Attn\in\mathbb{R}^{H_i\times W_i\times d_v}$ is the result of the scaled dot product with softmax normalization between $Q, K$ and $V$ with dimension $d_q, d_k$ equal to 16 and $d_v$ configured to $0.215 * C_i$. Finally, the output of RepSA is the projection with linear weight $W_o\in\mathbb{R}^{(C_i+d_v)\times C_i}$ to aggregate the local feature ($U$) and the global attention map ($Attn$).

\begin{table}[ht]
\begin{center}
\caption{All Variant RAViT Model configurations. \#Blocks denotes the number of RAViT blocks.}
\begin{tabular}{cccccccc}
\hline
\multicolumn{1}{c|}{\multirow{2}{*}{Stage}} & \multicolumn{1}{c|}{\multirow{2}{*}{Res}} & \multicolumn{1}{c|}{\multirow{2}{*}{Layer Spec}} & \multicolumn{4}{c}{RAViT}  \\ 
\cline{4-7} 
\multicolumn{1}{l|}{} & \multicolumn{1}{l|}{} & \multicolumn{1}{l|}{} & \multicolumn{1}{c|}{T26} & \multicolumn{1}{c|}{S22} & \multicolumn{1}{c|}{S26} & \multicolumn{1}{c}{M26} \\ 
\hline
\multicolumn{1}{l|}{\multirow{3}{*}{Stem}} & \multicolumn{1}{l|}{\multirow{3}{*}{$H\times W$}} & \multicolumn{1}{c|}{\multirow{3}{*}{\begin{tabular}[c]{@{}c@{}}Conv\\ Dims ($C_i$)\end{tabular}}} & \multicolumn{4}{c}{ $3 \times 3$, Stride 2} \\ 
\multicolumn{1}{l|}{} & \multicolumn{1}{l|}{} & \multicolumn{1}{l|}{}  & \multicolumn{4}{c}{ $3 \times 3$, Stride 2}\\
\cline{4-7}  
\multicolumn{1}{l|}{} & \multicolumn{1}{l|}{} & \multicolumn{1}{c|}{} & \multicolumn{1}{c|}{40} & \multicolumn{1}{c|}{48}  & \multicolumn{1}{c|}{48} & \multicolumn{1}{c}{64}  \\ 
\hline
\multicolumn{1}{l|}{\multirow{3}{*}{1}} & \multicolumn{1}{l|}{\multirow{3}{*}{$\frac{H}{4}\times\frac{W}{4}$}} & \multicolumn{1}{l|}{Mixer}  & \multicolumn{4}{c}{ RepMSDW $3 \times 3$ }  \\
\cline{3-7}
\multicolumn{1}{l|}{} & \multicolumn{1}{l|}{} & \multicolumn{1}{l|}{FFN ($r$)}  & \multicolumn{1}{c|}{3} & \multicolumn{1}{c|}{3} & \multicolumn{1}{c|}{3} & \multicolumn{1}{c}{3}   \\ 
\cline{3-7}
\multicolumn{1}{l|}{} & \multicolumn{1}{l|}{} & \multicolumn{1}{l|}{\#Blocks} & \multicolumn{1}{c|}{2}  & \multicolumn{1}{c|}{2}  & \multicolumn{1}{c|}{2} & \multicolumn{1}{c}{2} \\ 
\hline
\multicolumn{1}{l|}{\multirow{5}{*}{2}} & \multicolumn{1}{l|}{\multirow{5}{*}{$\frac{H}{8}\times\frac{W}{8}$}} & \multicolumn{1}{l|}{Downsample} & \multicolumn{4}{c}{RepMSDW $7 \times 7$, Stride 2} \\ 
\cline{4-7} 
\multicolumn{1}{l|}{} & \multicolumn{1}{l|}{} & \multicolumn{1}{l|}{Dims ($C_i$)} & \multicolumn{1}{c|}{80} & \multicolumn{1}{c|}{96}  & \multicolumn{1}{c|}{96} & \multicolumn{1}{c}{128}  \\ 
\cline{3-7} 
\multicolumn{1}{l|}{} & \multicolumn{1}{l|}{} & \multicolumn{1}{l|}{Mixer}  & \multicolumn{4}{c}{ RepMSDW $3 \times 3$ }  \\
\cline{3-7}
\multicolumn{1}{l|}{} & \multicolumn{1}{l|}{} & \multicolumn{1}{l|}{FFN ($r$)}  & \multicolumn{1}{c|}{3} & \multicolumn{1}{c|}{3} & \multicolumn{1}{c|}{3} & \multicolumn{1}{c}{3}   \\ 
\cline{3-7} 
\multicolumn{1}{l|}{} & \multicolumn{1}{l|}{} & \multicolumn{1}{l|}{\#Blocks} & \multicolumn{1}{c|}{4}  & \multicolumn{1}{c|}{4}  & \multicolumn{1}{c|}{4} & \multicolumn{1}{c}{4} \\ 
\hline
\multicolumn{1}{l|}{\multirow{5}{*}{3}} & \multicolumn{1}{l|}{\multirow{5}{*}{$\frac{H}{16}\times\frac{W}{16}$}} & \multicolumn{1}{l|}{Downsample} & \multicolumn{4}{c}{RepMSDW $7 \times 7$, Stride 2} \\ 
\cline{4-7} 
\multicolumn{1}{l|}{} & \multicolumn{1}{l|}{} & \multicolumn{1}{l|}{Dims ($C_i$)} & \multicolumn{1}{c|}{120} & \multicolumn{1}{c|}{192}  & \multicolumn{1}{c|}{192} & \multicolumn{1}{c}{256}  \\ 
\cline{3-7} 
\multicolumn{1}{l|}{} & \multicolumn{1}{l|}{} & \multicolumn{1}{l|}{Mixer}  & \multicolumn{4}{c}{ RepMSDW $7 \times 7$ }  \\
\cline{3-7}
\multicolumn{1}{l|}{} & \multicolumn{1}{l|}{} & \multicolumn{1}{l|}{FFN ($r$)}  & \multicolumn{1}{c|}{3} & \multicolumn{1}{c|}{3} & \multicolumn{1}{c|}{3} & \multicolumn{1}{c}{3}   \\ 
\cline{3-7}
\multicolumn{1}{l|}{} & \multicolumn{1}{l|}{} & \multicolumn{1}{l|}{\#Blocks} & \multicolumn{1}{c|}{16}  & \multicolumn{1}{c|}{12}  & \multicolumn{1}{c|}{16} & \multicolumn{1}{c}{16} \\ 
\hline
\multicolumn{1}{l|}{\multirow{5}{*}{4}} & \multicolumn{1}{l|}{\multirow{5}{*}{$\frac{H}{32}\times\frac{W}{32}$}} & \multicolumn{1}{l|}{Downsample} & \multicolumn{4}{c}{RepMSDW $7 \times 7$, Stride 2} \\ 
\cline{4-7} 
\multicolumn{1}{l|}{} & \multicolumn{1}{l|}{} & \multicolumn{1}{l|}{Dims ($C_i$)} & \multicolumn{1}{c|}{320} & \multicolumn{1}{c|}{384}  & \multicolumn{1}{c|}{384} & \multicolumn{1}{c}{512}  \\ 
\cline{3-7} 
\multicolumn{1}{l|}{} & \multicolumn{1}{l|}{} & \multicolumn{1}{l|}{Mixer}  & \multicolumn{4}{c}{ RepSA $7 \times 7$ }  \\
\cline{3-7}
\multicolumn{1}{l|}{} & \multicolumn{1}{l|}{} & \multicolumn{1}{l|}{FFN ($r$)}  & \multicolumn{1}{c|}{3} & \multicolumn{1}{c|}{3} & \multicolumn{1}{c|}{3} & \multicolumn{1}{c}{3}   \\ 
\cline{3-7} 
\multicolumn{1}{l|}{} & \multicolumn{1}{l|}{} & \multicolumn{1}{l|}{\#Blocks} & \multicolumn{1}{c|}{4}  & \multicolumn{1}{c|}{4}  & \multicolumn{1}{c|}{4} & \multicolumn{1}{c}{4} \\ 
\hline
\multicolumn{3}{c|}{Classifier Head} & \multicolumn{4}{c}{Avg Pool, FC (1280) }\\ \hline
\end{tabular}
\label{tab:arch_variant}
\end{center}
\end{table}

\subsection{Fast-FCOS}
We enhanced the FCOS by employing the proposed RAViT as the backbone for fast object detection in driving scenes. Additionally, we introduced a Reparameterized Feature Pyramid Network (FPN) for multi-scale feature extraction within the neck section of FCOS. As illustrated in Fig. \ref{fig:RepFPN}, we utilized three feature levels from the RAViT backbone, $F3, F4,$ and $F5$, and substituted the original $3\times3$ convolution of the typical FCOS FPN with a sequence of RepMSDW and $1\times1$ convolution after the aggregation of two scale features. RepMSDW is capable of extracting spatial features at different scales due to its multiple kernel scales, which can be reparameterized for a faster inference phase. 

In contrast to the original FCOS FPN that extends the three feature levels $\{F3, F4, F5\}$ from the backbone into five levels $\{P3, P4, P5, P6, P7\}$ with stride factors $\{8, 16, 32, 64, 128\}$, we employ only three feature levels $\{P3, P4, P5\}$ in a shared head for the classification of objects, center-ness, and bounding box regression. Since we use only three levels of features, the regression range is configured as 0, 128, 256, and 512 in the regression head. Due to the different sizes of feature levels, the regression range is adjusted with a different range for each level. The regression range of $P3$ is $\{0,128\}$, $\{128,256\}$ for $P4$ and $\{256,512\}$ for $P5$.    

\begin{table*}[!ht]
\centering
\caption{Comparison of RAViT variant with State-Of-The-Art on ImageNet-1K Dataset. GPU inference throughput is measured using RTX3090. NPU and Edge are the latency that was measured on the iPhone 15 Pro using CoreML format and the NVIDIA Jetson Orin Nano Edge Device using ONNX format.}
\begin{tabular}{ m{3.8cm}|>{\centering}m{0.8cm}|>{\centering}m{1.1cm}|>{\centering}m{1.0cm}|>{\centering}m{1.1cm}|>{\centering}m{1.0cm}|>{\centering}m{0.8cm}|>{\centering}m{0.8cm}|c }
\hline
\multirow{2}{*}{Model} & \multirow{2}{*}{Type} & Img  & Param & FLOPs & GPU & NPU & Edge & \multirow{2}{*}{Top-1} \\ 
                          & & Size  &   (M)  & (G)  & (Img/s) & (ms) &  (ms) &   \\ \hline
EfficientFormerV2-S1\cite{li2023rethinking} &Hybrid & 224 & 16.1 & 0.7 & 1153 & 0.76 & 14.3 & 77.9 \\
MobileViTV2-1.0\cite{mehta2023separable}    &Hybrid & 224 & 4.9  & 1.8 & 1858 & 1.94 & 14.4 & 78.1 \\
MobileOne-S3\cite{vasu2023mobileone}        &Conv   & 224 & 10.1 & 1.8 & 2793 & 0.93 & 15.2 & 78.1 \\
\rowcolor{gray!30}
\textbf{RAViT-T26}                    &Hybrid & 224 & 8.2  & 0.9 & 3755 & 0.71 & 12.9 & 78.4 \\
\hline 
EMO-6M\cite{zhang2023rethinking}      &Hybrid & 224 & 5.1  & 1.0 & 1916 & 2.49 & 16.9 & 79.0 \\
FastViT-T12\cite{vasu2023fastvit}     &Hybrid & 256 & 6.8  & 1.4 & 3182 & 0.99 & 10.9 & 79.1 \\
EdgeNeXt-S \cite{maaz2022edgenext}    &Hybrid & 256 & 5.6  & 1.3 & 1207 & 32.5 & 13.1 & 79.4 \\
MobileOne-S4 \cite{vasu2023mobileone} &Conv   & 224 & 14.8 & 2.9 & 1892 & 1.21 & 19.0 & 79.4 \\
RepViT-M1.1\cite{wang2024repvit}      &Conv   & 224 & 8.2  & 1.3 & 3604 & 0.74 & 11.4 & 79.4 \\
\rowcolor{gray!30}
\textbf{RAViT-S22}                &Hybrid & 224 & 10.2 & 1.2 & 3491 & 0.80 & 11.9 & 79.6 \\
\hline
FastViT-S12\cite{vasu2023fastvit} &Hybrid & 256 & 8.8  & 1.8 & 2313 & 1.13 & 11.8 & 79.8 \\
PoolFormer-S24\cite{yu2022metaformer} &Hybrid & 224 & 21.4 & 3.4 & 1424 & 1.68 & 18.3 & 80.3 \\
MobileViTV2-1.5 \cite{mehta2023separable} &Hybrid & 256 & 10.6 & 4.2 & 1116 & 2.95 & 20.9 & 80.4 \\
EfficientFormerV2-S2\cite{li2023rethinking} &Hybrid & 224 & 12.6 & 1.3 & 611  & 1.23 & 21.5 & 80.4 \\
\rowcolor{gray!30}
\textbf{RAViT-S26}                &Hybrid & 224 & 11.5 & 1.4 & 3079 & 0.83 & 13.0 & 80.2 \\
\rowcolor{gray!30}
\textbf{RAViT-SA22}               &Hybrid & 224 & 10.9 & 1.3 & 2878 & 0.94 & 17.3 & 80.4 \\

\hline
FastViT-SA12\cite{vasu2023fastvit}     &Hybrid & 256 & 10.9 & 1.9 & 2181 & 1.22 & 12.8 & 80.6 \\
RepViT-M1.5 \cite{wang2024repvit}      &Conv   & 224 & 14.0 & 2.3 & 2151 & 1.04 & 19.4 & 81.2 \\
Swin-T \cite{liu2021swin}              &ViT    & 224 & 29.0 & 4.5 & 886  & 6.51 & 24.5 & 81.3 \\
PoolFormer-S36\cite{yu2022metaformer} &Hybrid & 224 & 30.9 & 5.0 & 967 & 2.35 & 26.8 & 81.4 \\
\rowcolor{gray!30}
\textbf{RAViT-M26}                &Hybrid & 224 & 18.5 & 2.4 & 2193 & 1.17 & 14.4 & 81.4 \\
\hline
\end{tabular}
\label{tab:imagenet}
\end{table*}

\begin{table*}[!ht]
\centering
\caption{Instance segmentation test of RAViT as a backbone on COCO val2017 with Mask R-CNN. $AP^b$ and $AP^m$ denote bounding box average precision and mask average precision, respectively. The FLOPs and latency are measured at resolution 1280 $\times$ 800.
}
\begin{tabular}{ m{2.8cm}|>{\centering}m{0.8cm}|>{\centering}m{0.8cm}|>{\centering}m{0.8cm}|>{\centering}m{0.8cm}|>{\centering}m{0.8cm}|>{\centering}m{0.8cm}|>{\centering}m{0.8cm}|>{\centering}m{0.8cm}|>{\centering}m{0.8cm}|c} \hline
\multirow{2}{*}{Backbone} & \multirow{2}{*}{$AP^{b}$}  & \multirow{2}{*}{$AP^{b}_{50}$} & \multirow{2}{*}{$AP^{b}_{75}$} & \multirow{2}{*}{$AP^{m}$}  & \multirow{2}{*}{$AP^{m}_{50}$} & \multirow{2}{*}{$AP^{m}_{75}$} & Param & FLOPs & GPU & EDGE \\
    &  &  &  &  &  &  & (M) & (G) & (img/s) & (ms)   \\ \hline

PVT-T\cite{wang2021pyramid} & 36.7 & 59.2 & 39.3 & 35.1 & 56.7 & 37.3 & 32.8 & 239.8  & 24.7 &  517 \\ 
PoolFormer-S12\cite{yu2022metaformer} & 37.3 & 59.0 & 40.1 & 34.6 & 55.8 & 36.9 & 31.6  & 207.3  & 20.0 & 353 \\
ResNet-50\cite{he2016deep}   & 38.0 & 58.6 & 41.4 & 34.4 & 55.1 & 36.7 & 44.2 & 260.1 & 31.5 & 333 \\
FastViT-SA12\cite{vasu2023fastvit} & 38.9 & 60.5 & 42.2 & 35.9 & 57.6 & 38.1 & 30.5 & 200.4 & 37.5 & 413 \\
RepViT-M1.1\cite{wang2024repvit} & 39.8 & 61.9 & 43.5 & 37.2 & 58.8 & 40.1 & 27.9 & 197.6 & 42.2 & 370 \\ 
\rowcolor{gray!30}
\textbf{REViT-S26}           & 40.4 & 62.5 & 44.2 & 37.8 & 59.8 & 40.2 & 29.6 & 198.8 & 40.0 & 284 \\ \hline
PoolFormer-S24\cite{yu2022metaformer} & 40.1 & 62.2 & 43.4 & 37.0 & 59.1 & 39.6 & 41.0 & 213.8 & 12.2 & 558 \\ 
PVT-S\cite{wang2021pyramid}  & 40.4 & 62.9 & 43.8 & 37.8 & 60.1 & 40.3 & 44.1 & 304.5 & 16.9 & 731 \\ 
RepViT-M1.1\cite{wang2024repvit} & 41.6 & 63.2 & 45.3 & 38.6 & 60.5 & 41.4 & 33.8 & 217.1 & 35.8 & 427 \\ 
\rowcolor{gray!30}
\textbf{REViT-M26}           & 41.6 & 64.0 & 45.3 & 38.9 & 60.9 & 41.9 & 34.4 & 212.3 & 36.7 & 334 \\ \hline

\end{tabular}
\label{tab:obj}
\end{table*}
\section{Result}
To evaluate the effectiveness of the RAViT backbone and Fast-COS, we perform several tests. The ImageNet-1K dataset with 1000 categories is selected as the image classification benchmarking for the backbone test. We also perform a backbone evaluation test for instance segmentation on the COCO dataset. In a specific task, we advance Fast-COS with RepFPN to perform driving scene object detection on the BDD100K and TJU-DHD-traffic datasets. Since the FLOPs do not directly affect the computational complexity, we compare RAViT and Fast-COS with other models in the state-of-the-art using throughput and latency metrics that were tested on three different wide-range application processing devices, including the GPU RTX3090, the Neural Processing Unit (NPU) on the iPhone 15 Pro as a mobile processing unit, and the Jetson Orin Nano as the edge device processing unit. This extensive benchmark will represent real-time performance on diverse hardware platforms. 
\subsection{Evaluation Result of RAViT Backbone}
\subsubsection{Setup}
We evaluate RAViT using ImageNet-1K \cite{deng2009imagenet} as the most popular image classification benchmarking dataset. ImageNet-1K has 1000 categories with 1.2 million images for training and 50000 images for validation. We follow a training recipe in \cite{liu2021swin} with a total of 300 epochs in each RAViT model variant, with a resolution of 224$\times$224. Data augmentation and regularization approaches encompass several methods such as RandAugment \cite{cubuk2020randaugment}, Mixup, CutMix \cite{yun2019cutmix}, Random Erasing \cite{zhong2020random}, weight decay, label smoothing, and Stochastic Depth. We employ the AdamW optimizer with a 0.004 base learning rate and a total batch size of 2046 on 4$\times$A6000 GPUs for most RAViT models.

We also conducted an experiment on the COCO dataset, a benchmark widely used for object detection and instance segmentation using Mask R-CNN. The backbone of Mask R-CNN was replaced with RAViT, which leverages multi-scale convolution and self-attention to improve feature extraction. The COCO dataset's training split was used for model training, while the validation split was used for evaluation, adhering to the standard COCO metrics, including mean Average Precision (mAP) for both bounding box detection and segmentation masks.

We assess performance through inference latency tests conducted on two types of constrained-resource device hardware and a single desktop GPU. For mobile device performance assessment, the iPhone 15 Pro is utilized. All models are transformed into the CoreML format, and each undergoes 50 $\times$ inference loops following a prior 20-loop warm-up. The mean inference duration serves as the evaluation metric. The Jetson Orin Nano is chosen to represent Edge device hardware for evaluation purposes. Models are adapted to the ONNX format for latency measurement on Edge devices. In Edge device assessment, a warm-up duration of 20 seconds precedes 1000 $\times$ inference loops.

\subsubsection{Benchmarking on ImageNet-1K}
The comparative analysis presented in Table \ref{tab:imagenet} highlights the performance of RAViT variants against state-of-the-art models on the ImageNet-1K dataset. We evaluate across various hardware platforms including GPU, Mobile NPU, and Edge devices to give a wide application illustration as a backbone. The RAViT models demonstrate competitive accuracy and computational efficiency trade-offs when compared to recent state-of-the-art models. For instance, RAViT-M26 reaches 81.4\% Top-1 accuracy while achieving $2.27\times$ faster GPU throughput, $2\times$ faster NPU latency, and $1.8\times$ faster Edge device latency compared to PoolFormer-S36 and Swin-T. RAViT-M26 achieves higher 0.2\% Top-1 accuracy while maintaining similar NPU and Edge device latency compared to the recent mobile vision transformer such as RepViT, showcasing its architectural efficiency. Similarly, RAViT-M26 achieves higher 0.8\% Top-1 accuracy compared to FastViT-SA12, while maintaining 4\% faster mobile NPU inference speed. 

The RAViT models consistently achieve high accuracy across different configurations. Although MobileOne also uses the reparameterization technique, the hybrid transformer architecture of RAViT indicates a trade-off between accuracy and speed. For example, RAViT-S22 can outperform MobileOne-S4 with $1.8\times$ faster GPU throughput, $1.5\times$ faster Mobile NPU latency, and $1.6\times$ faster Edge device latency. RAViT models achieve faster GPU, Mobile NPU, Edge device inference while providing superior accuracy, showcasing their versatility and effectiveness for real-world deployment scenarios.
\subsection{Benchmarking with SOTA Models in COCO Instance Segmentation}
Table III presents the evaluation of the proposed RAViT backbones on the COCO val2017 dataset with Mask R-CNN, compared to other state-of-the-art methods. The evaluation metrics include bounding box average precision $(AP_b)$, mask average precision $(AP_m)$, computational complexity (FLOPs), parameter size, and latency on GPU and EDGE devices. The results highlight the ability of RAViT to achieve a favorable balance between segmentation accuracy and computational efficiency.

The RAViT backbones demonstrate competitive or superior performance in both bounding box and mask precision. RAViT-S26 achieves an $AP_b$ of 40.4 and an $AP_m$ of 37.2, outperforming PVT-S and PoolFormer-S24. RAViT-M26 achieves the highest scores among all methods, with 41.6\% of $AP_b$   and 38.9\% of $AP_m$, surpassing RepViT-M1.1 while only using the convolution mixer in accuracy, while maintaining comparable efficiency. These results validate the effectiveness of the RAViT architecture in improving instance segmentation performance.

Latency and inference speed further highlight the efficiency of the RAViT architecture. RAViT-S26 achieves the highest GPU inference speed of 40.0 images per second, significantly outperforming PVT-S $2.4\times$ and 8.25\% faster than RepViT-M1.1. On EDGE devices, RAViT-S26 achieves the lowest latency at 284 ms, making it highly suitable for real-time applications. RAViT-M26 also delivers competitive EDGE latency at 334 ms, outperforming several other methods in the comparison.

Compared to PoolFormer, PVT, and RepViT, the proposed RAViT architecture strikes an excellent balance between accuracy and efficiency. RAViT-M26 achieves the highest accuracy across the evaluated models, while RAViT-S26 stands out for its minimal latency and high inference speed. These results demonstrate the scalability of RAViT, offering lightweight and high-performance variants that cater to diverse deployment scenarios. In conclusion, the proposed RAViT backbones are well-suited for both performance-driven and latency-critical applications, solidifying their value in instance segmentation tasks.

\begin{table*}[!ht]
\centering
\caption{Object detection on BDD100K and TJU-DHD-Traffic dataset with  FCOS-RAViT. $AP^b$ denote bounding box average precision. The FLOPs, GPU and EDGE inference throughput are measured at resolution 1280 $\times$ 720.
}
\begin{tabular}{ c|m{1.9cm}|m{1.9cm}|>{\centering}m{0.7cm}|>{\centering}m{0.8cm}|>{\centering}m{0.8cm}|>{\centering}m{0.7cm}|>{\centering}m{0.7cm}|>{\centering}m{0.7cm}|>{\centering}m{0.7cm}|>{\centering}m{0.8cm}|>{\centering}m{0.8cm}|c} \hline
\multirow{2}{*}{Dataset}&\multirow{2}{*}{Network} & \multirow{2}{*}{Backbone} & \multirow{2}{*}{$AP$} & \multirow{2}{*}{$AP_{50}$}& \multirow{2}{*}{$AP_{75}$} & \multirow{2}{*}{$AP_{s}$}  & \multirow{2}{*}{$AP_{m}$} & \multirow{2}{*}{$AP_{l}$} & Param & FLOPs & GPU & EDGE \\
           & &  & & &  &  &  &  & (M) & (G) & (img/s) & (img/s)   \\ \hline
\multirow{6}{*}{BDD100K}&YOLOF\cite{mboutayeb2024fcosh}&ResNet-50\cite{he2016deep}& 24.5 & 45.1 & 22.9 & 6.6 & 30.3 & 51.0 & 42.5 & 94.5& 60.7 & 9.9  \\ 
& FCOS\cite{mboutayeb2024fcosh}&ResNet-50\cite{he2016deep}
                        &29.0 &52.9 &26.8 &12.3 &34.7 &50.6 &31.9 &181.6 &39.3 & 6.8  \\
&FCOS\cite{mboutayeb2024fcosh}&ResNet-101\cite{he2016deep} 
                        &30.0 &54.2 &27.9 &12.6 &35.8 &52.2 &50.8 &251.3 &28.2 &-\\ \cline{2-13}
& \multirow{2}{*}{FCOS\cite{9010746}}&RAViT-S26 &30.2 &54.6 &28.3 &12.7 &35.2 &52.3 &17.4 &132.6 &51.8 & 8.6 \\
                    &  &RAViT-M26 &30.5 &55.3 &28.5 &13.2 &35.1 &53.2 &24.6 &152.8 &45.7 & 7.6 \\ \cline{2-13}
&\multirow{2}{*}{Fast-COS} &RAViT-S26 &31.1 &56.6 &29.3 &14.0 &36.5 &51.7 &14.7 &120.9 &57.0 & 9.3 \\
                    &  & RAViT-M26 &31.8 &57.2 &30.1 &14.4 &37.2 &53.6 &21.9 &141.2 &49.6 & 8.1 \\
\hline
\multirow{6}{*}{TJU-DHD}&$^\dagger$RetinaNet\cite{ross2017focal}&ResNet-50\cite{he2016deep} & 53.5 & 80.9 & 60.0 & 24.0 & 50.5 & 68.0 & 36.4 & 216.1 & 33.7 & 6.9 \\ 
&$^\dagger$FCOS\cite{9010746}&ResNet-50\cite{he2016deep}  
                        &53.8 &80.0 &60.1 &24.6 &50.6 &50.6 &32.1 &211.6 &33.0 &  6.9  \\ \cline{2-13}
&\multirow{2}{*}{$^\dagger$FCOS\cite{9010746}}& RAViT-S26 &53.9 &79.3 &61.0 &26.2 &50.4 &69.1 &17.4 &143.9 &48.5 &8.6 \\
                      & & RAViT-M26 &54.5 &80.0 &61.4 &27.2 &50.7 &70.0 &24.6 &165.9 &47.7 &7.6 \\
\cline{2-13} 
&\multirow{2}{*}{Fast-COS} &RAViT-S26 &53.4 &79.2 &60.3 &26.1 &49.7 &68.3 &17.4 &120.9 &56.8 & 9.3 \\
                    & & RAViT-M26 &54.5 &80.0 &61.4 &26.9 &50.5 &69.4 &21.9 &141.2 &49.4 &8.1 \\
                      
\hline
\end{tabular}
\label{tab:driver-scene}
\end{table*}

\subsection{Evaluation Result of Fast-COS on driving scene Object Detection Task}
\subsubsection{Setup} The proposed model was tested on two high-resolution, large-scale datasets: BDD100K and TJU-DHD-traffic, both suited for assessing detection networks from a driver's viewpoint. These datasets include diverse scenes such as urban streets and residential areas, with 1.84 million and 239,980 annotated bounding boxes, respectively. BDD100K features 10 categories, including Bus and Car, while TJU-DHD covers 5 categories like Pedestrian and Cyclist, available under different lighting conditions. Both datasets offer scenarios in various weather conditions, providing a valuable resource for real-world model testing. Organization comprises 45,266 training images and 5,000 validation images for TJU-DHD Traffic, and 70,000 training images and 10,000 validation images for BDD100K.

In the driving scene object detection experiment, we use an input size of 1280 × 720 pixels in both training and evaluation. Training utilizes 4 NVIDIA GPUs, each handling a mini-batch of 8 images. The AdamW optimizer governs training, starting with a learning rate of 0.0001, which is reduced by a factor of 10 at the 8th and 11th epochs. The model undergoes end-to-end training with RAViT, initialized from pre-trained weights for efficient learning. Data augmentation, including random flipping and resizing, is applied for robustness. During inference, the top 100 detection bounding boxes per image are recorded for performance assessment.
\subsubsection{Benchmarking on BDD100K and TJU-DHD Traffic}
A comparative study illustrated in Table \ref{tab:driver-scene} showcases the performance of Fast-COS utilizing RAViT backbone variants, in comparison with original FCOS models, evaluated on the BDD100K and TJU-DHD Traffic datasets. The findings demonstrate that the RAViT backbone can enhance the inference speed of FCOS by 62\% over the original FCOS-ResNet-101, when juxtaposed with FCOS-RAViT-M26. Additionally, the RAViT backbone contributes to an increase in prediction accuracy, indicated by a 2\% improvement in $AP_{50}$ in the BDD100K dataset test. By integrating the RAViT-M26 model with the RepFPN, which includes the use of RepMSDW, the Fast-COS can achieve $AP_{50}$ accuracy 5.5\% higher along with a 75.9\% improvement in GPU inference speed when compared to the FCOS-ResNet-101. 

In the TJU-DHD Traffic dataset, employing the RAViT-M26 as a backbone enhances the overall $AP$ by 1.3\%, with a 38\% enhancement in predicting large objects ($AP_l$). Utilizing the same input size configuration ($1333\times800$), the proposed RAViT backbone also increases detection speed by 44.5\% compared to the original FCOS. To achieve a detection speed comparable to the BDD100K test, the Fast-COS model variant was evaluated with an input size of $1280\times720$ pixels. In this configuration, Fast-COS boosts detection speed by 49.6\% relative to the original FCOS, while maintaining similar prediction accuracy to the FCOS-RAViT variant trained with a larger input size.

In the evaluation conducted on the Edge device utilizing Jetson Orin Nano with the ONNX framework, the peak performance of the Fast-COS variant is observed when employing RAViT-S26, achieving a prediction speed of 9.3 FPS across both driving scene dataset tests. Despite being 6.5\% slower than YOLOF-ResNet50, Fast-COS with RAViT-S26 attains a 26.9\% improvement in overall average precision $AP$ in BDD100K test. The comprehensive hardware assessment indicates that while several models can exceed 30 FPS for GPU inference speed, optimization on the EDGE device is essential to attain at least 15 FPS for it to be viable as a real-time driving scene object detection hardware option.

\begin{figure*}
\centering
         \includegraphics[width=5.8cm]{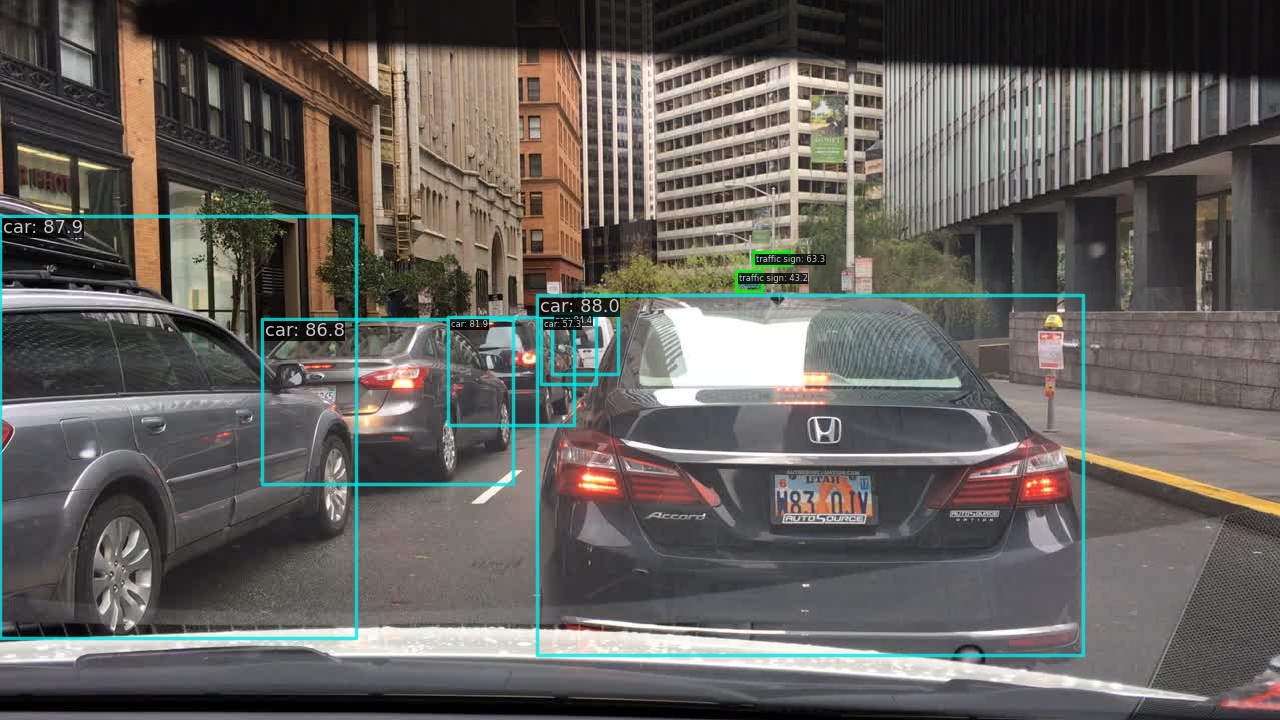}
         \vspace{0.25mm}
         \includegraphics[width=5.8cm]{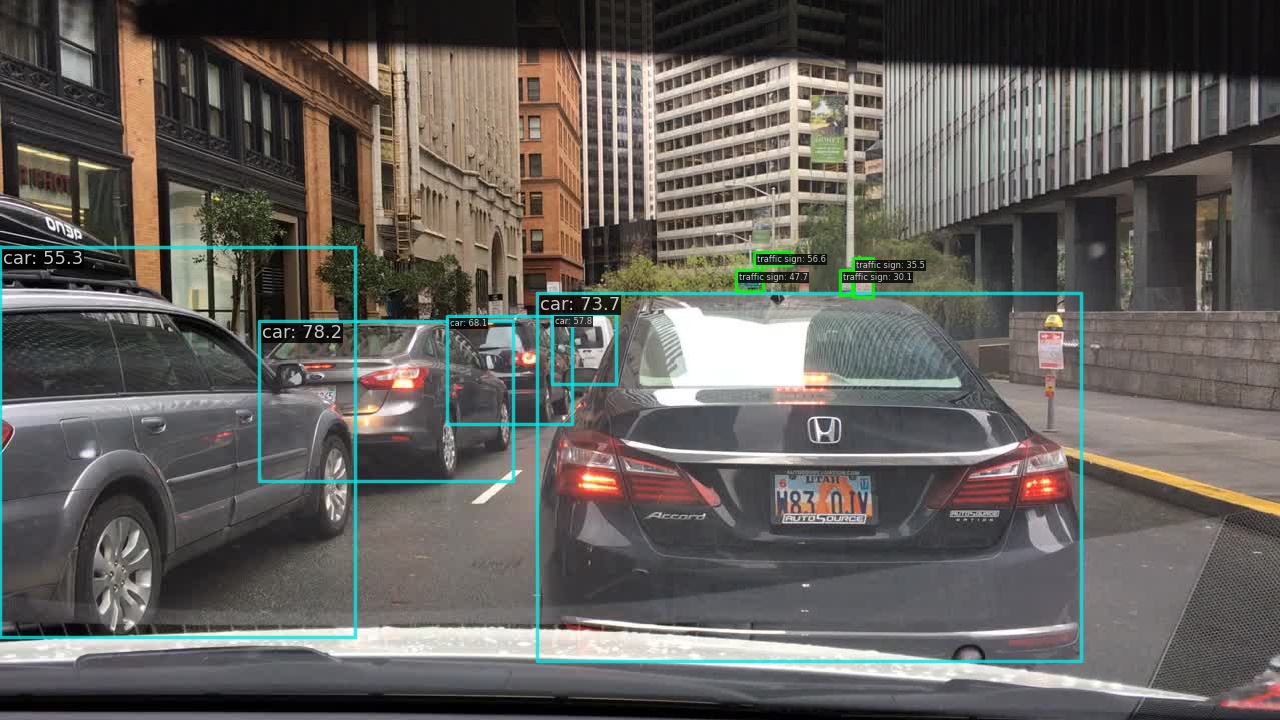}
         \vspace{0.25mm}
         \includegraphics[width=5.8cm]{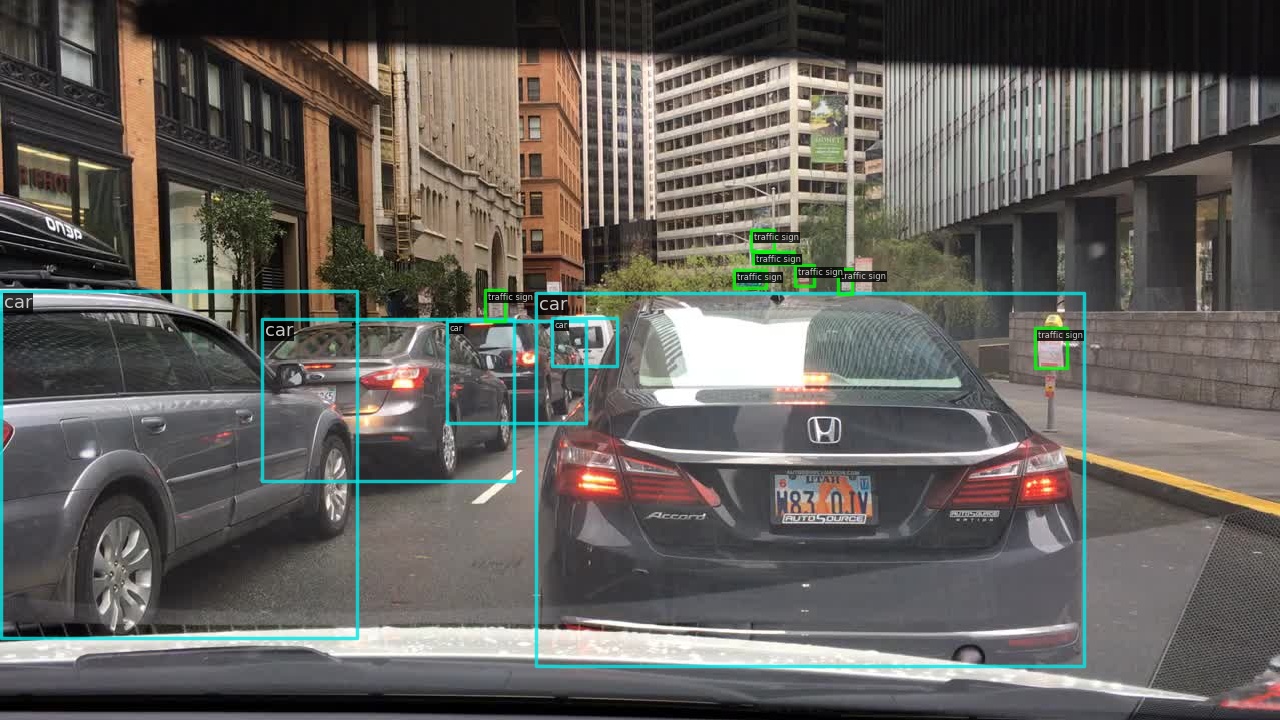}
         \vspace{0.25mm}
         \includegraphics[width=5.8cm]{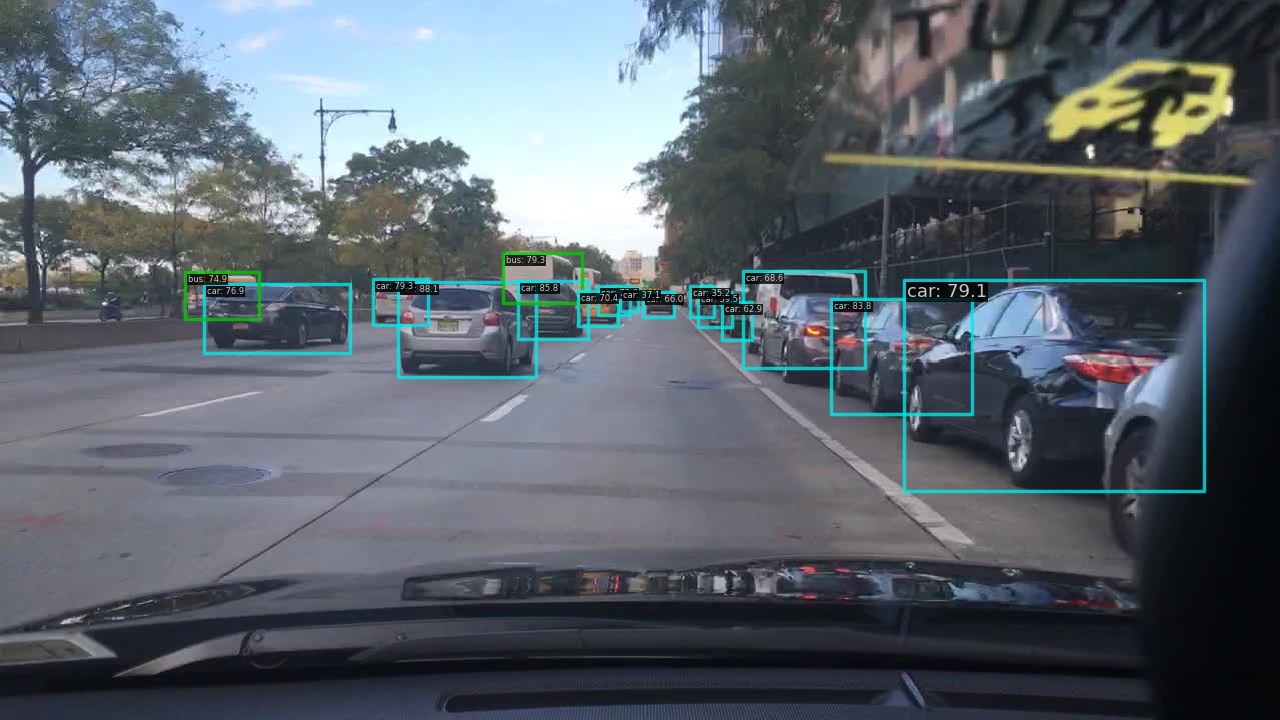}
         \vspace{0.25mm}
         \includegraphics[width=5.8cm]{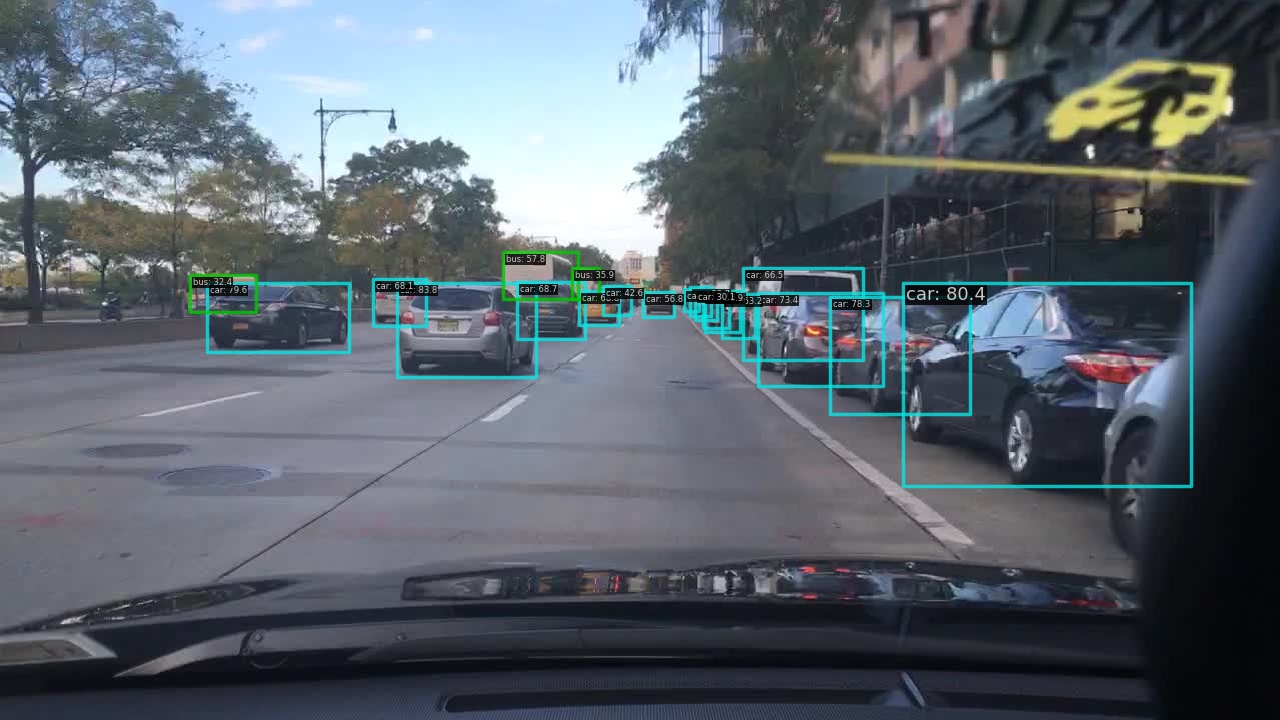}
         \vspace{0.25mm}
         \includegraphics[width=5.8cm]{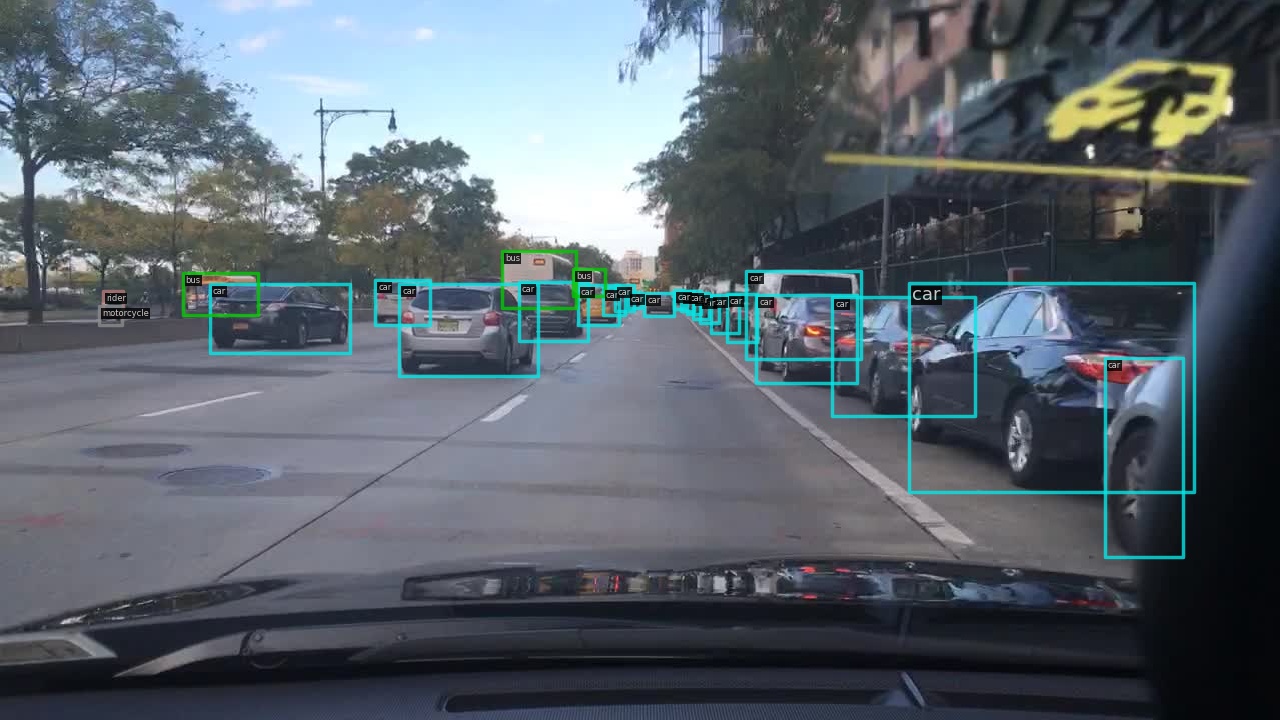}
         \vspace{0.25mm}
         \includegraphics[width=5.8cm]{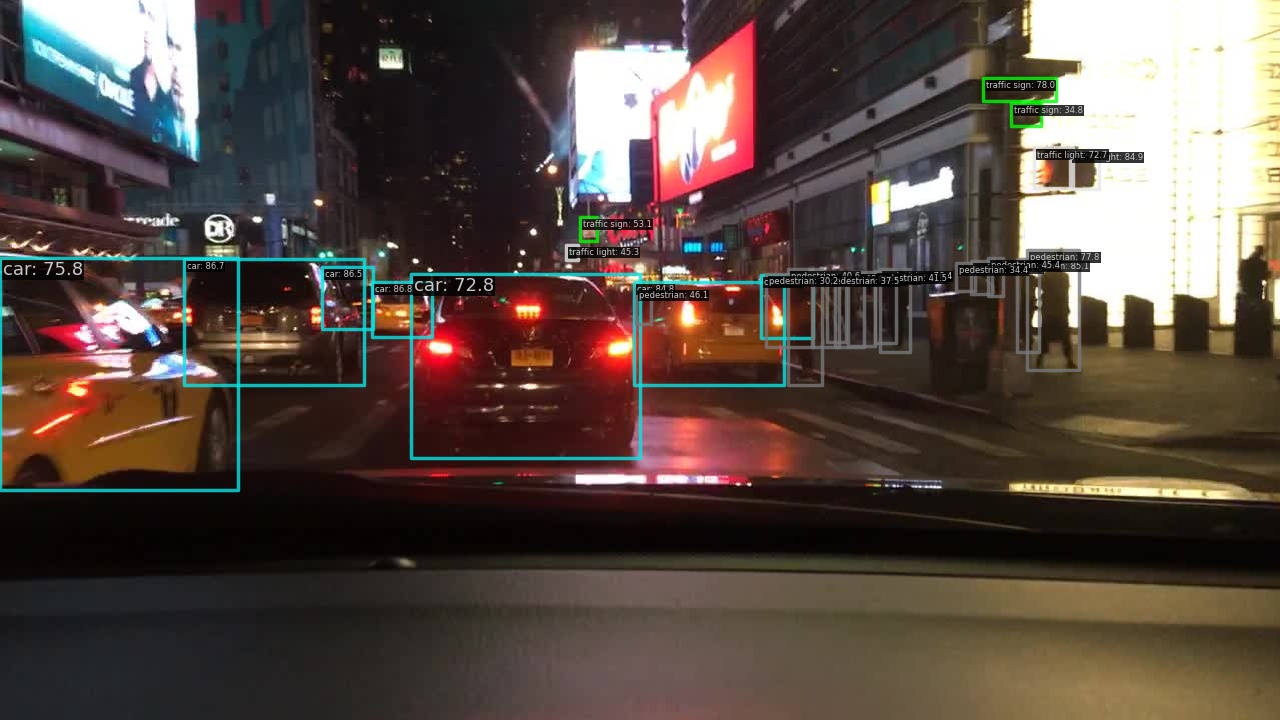}
         \vspace{0.25mm}
         \includegraphics[width=5.8cm]{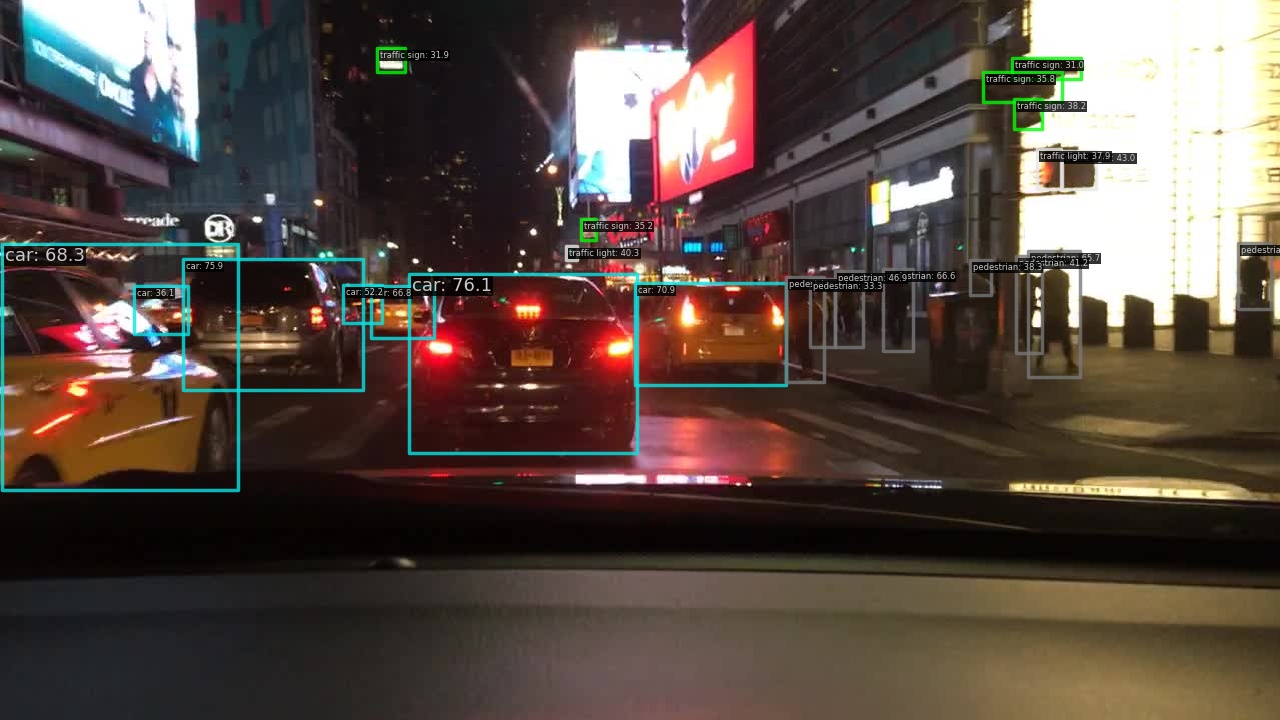}
         \vspace{0.25mm}
         \includegraphics[width=5.8cm]{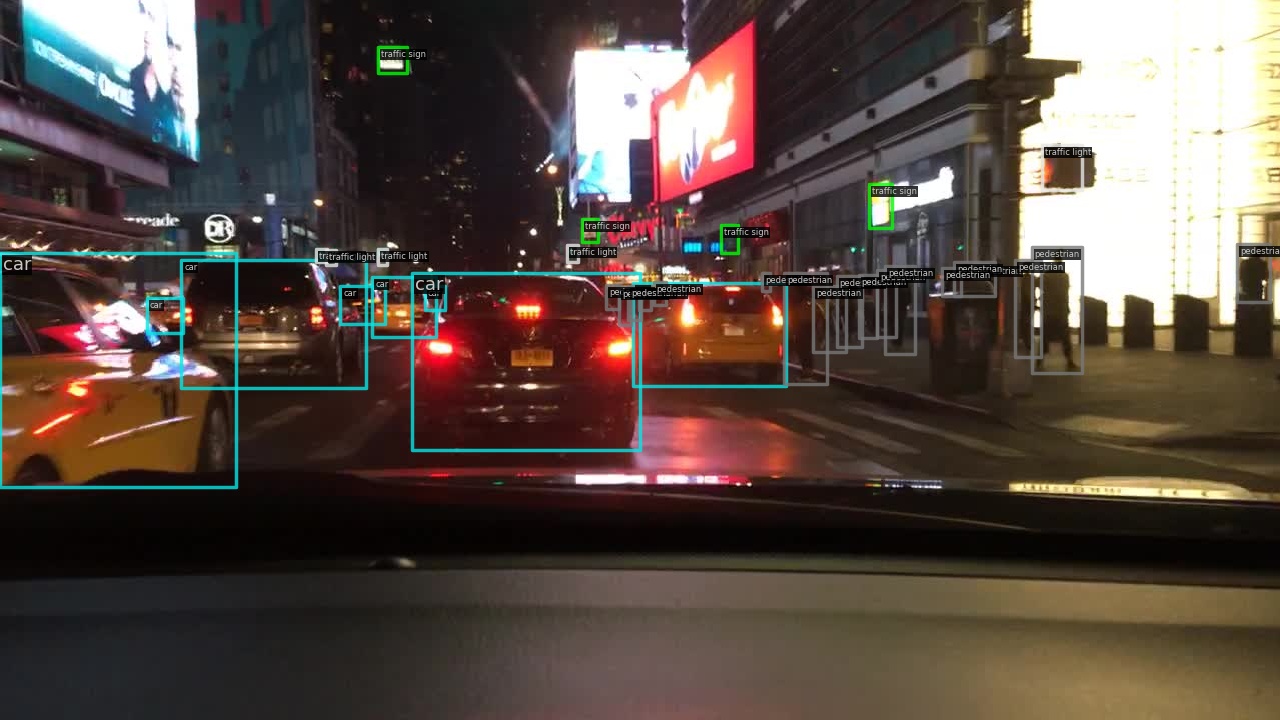}
         \vspace{0.5mm}
         \includegraphics[width=5.8cm]{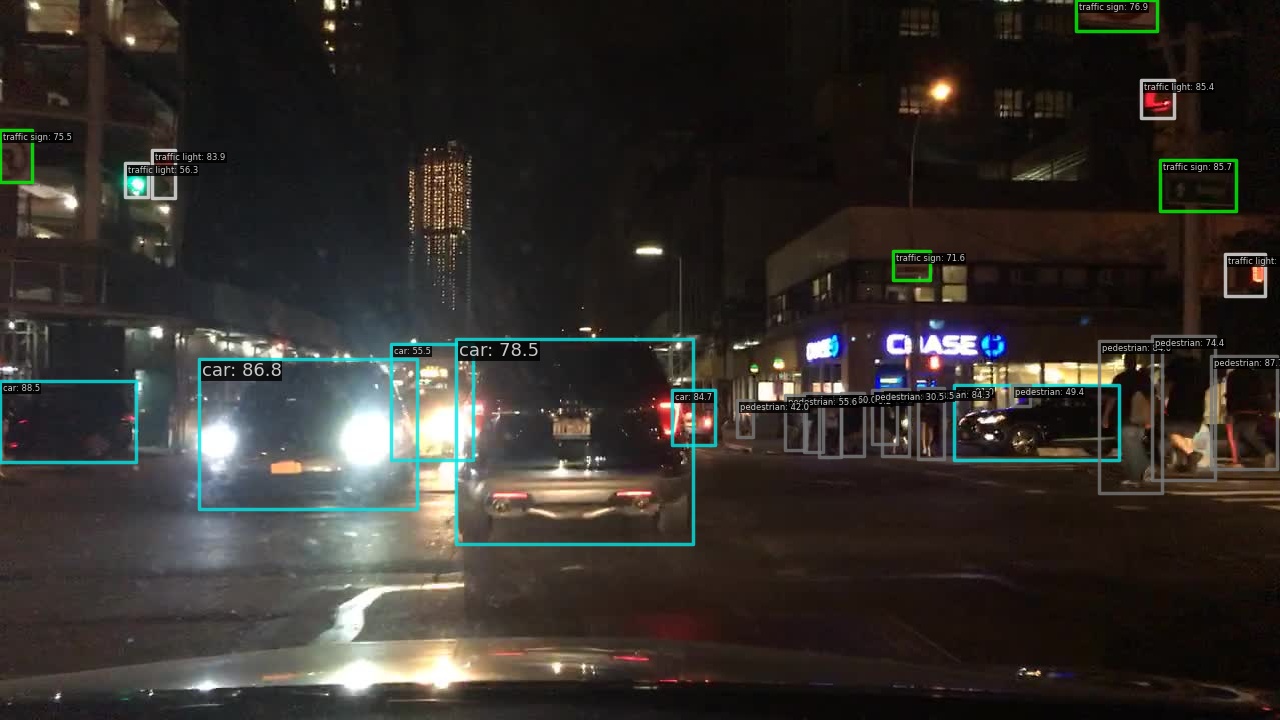}
         \includegraphics[width=5.8cm]{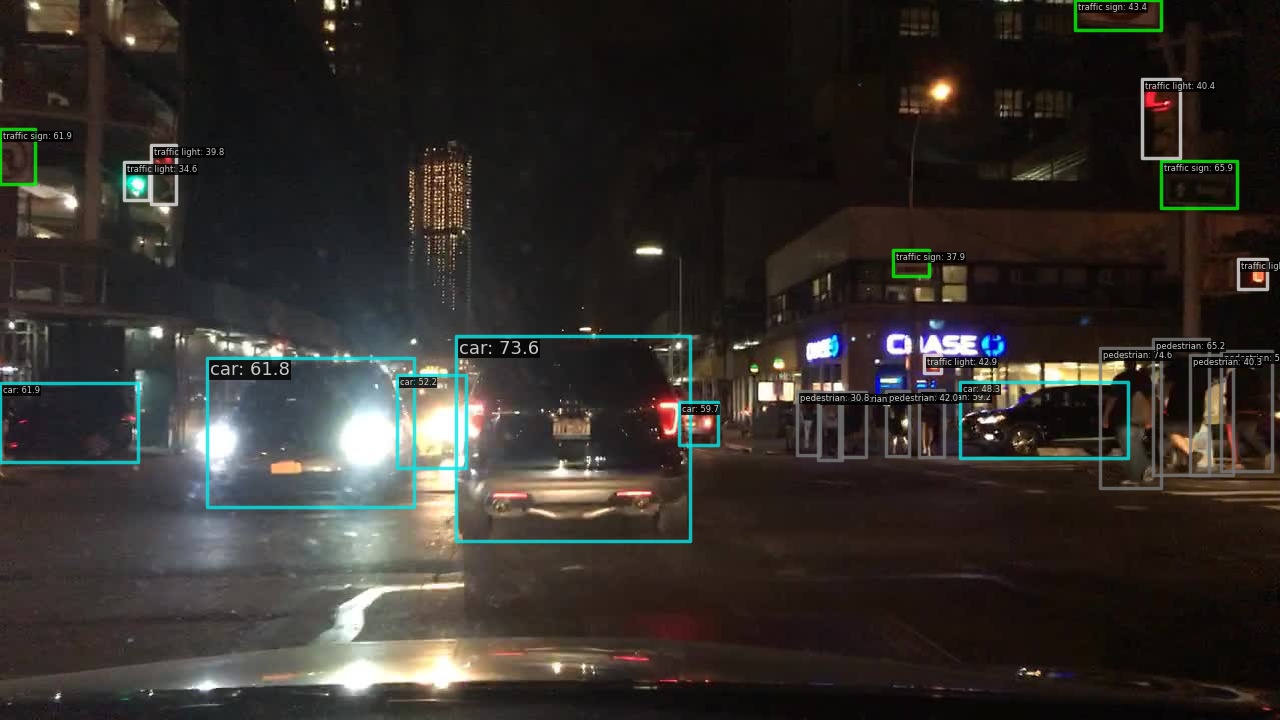}
         \includegraphics[width=5.8cm]{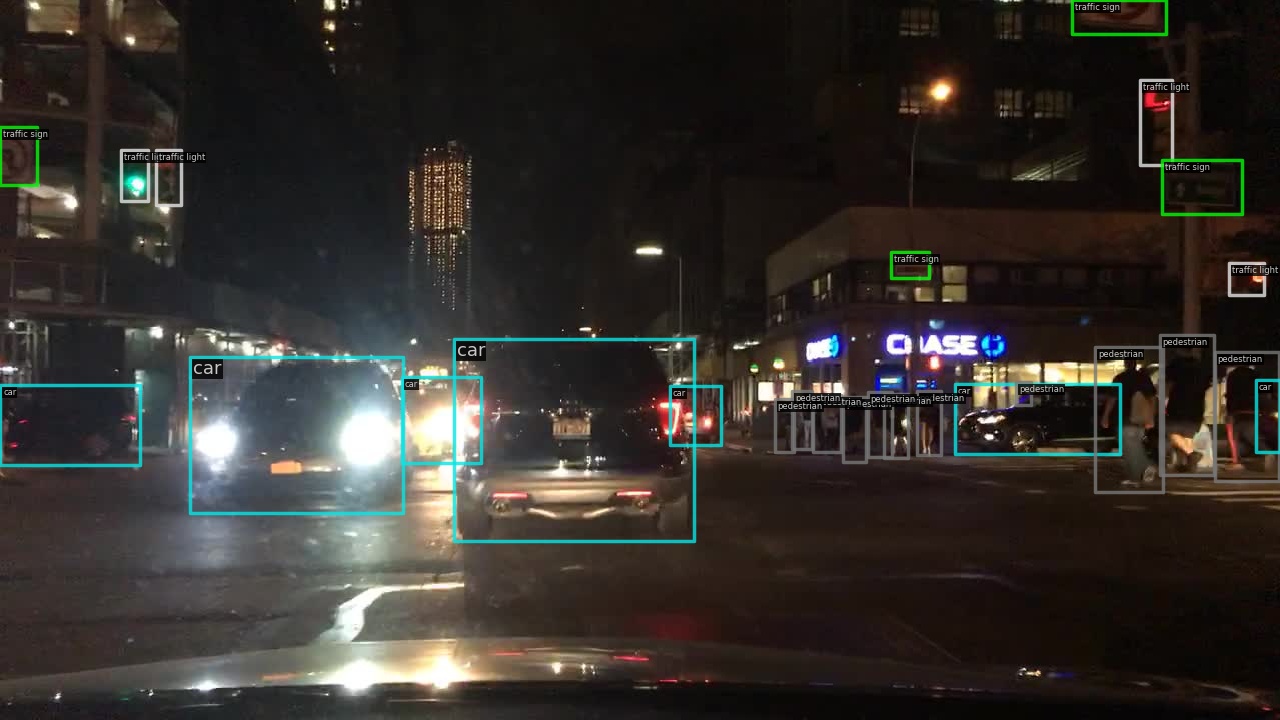}
         \\(1)\hspace{5.8cm}(2)\hspace{5.8cm}(3)
\caption{The detection results using (1). FCOS+ResNet-50 and (2). Fast-COS+RAViT-M26 compare to (3). ground truth}
\label{fig:qualitativeobj}
\end{figure*}

\subsection{Ablation Experiment}
\subsubsection{Multi-Scale in RepMSDW and Combination with Self-Attention}
We conducted an ablation study on Multi-Scale convolution kernel sizes and compared them against sole square kernel reparameterization and configurations without kernel reparameterization. This analysis was specifically performed using the RAViT-S22 variant. As depicted in Table \ref{tab:ablation_token_mixer}, adopting Multi-Scale kernel sizes during reparameterization results in enhanced classification accuracy on the ImageNet-1K dataset. Utilizing multi-scale reparameterization offers a 0.12\% improvement over single square reparameterization, as referenced in \cite{wang2024repvit, vasu2023fastvit, ding2021repvgg}, and a 0.22\% gain over configurations without reparameterization. 

Given that RepMSDW is constrained in capturing long dependencies, we evaluated its integration with self-attention, a technique employed in Transformer models \cite{li2022efficientformer, liu2023efficientvit, liu2021swin}. Initially, we incorporated RepMSDW in Multi-Head Self-Attention (MHSA) to substitute the $3\times3$ DWConv, as described in \cite{li2023rethinking}, with a $7\times7$ RepMSDW. This modification increased accuracy to 79.2\%. However, MHSA requires numerous array transformations, impacting GPU throughput and NPU latency. Subsequently, we adopted single-head attention (SA), as proposed in \cite{yun2024shvit}, to address the computational redundancies in MHSA. Combining RepMSDW with SA elevated accuracy to 79.6\% while maintaining efficiency in both GPU throughput and NPU latency.
\begin{table}[!ht]
\centering
\caption{Ablation of RepMSDW and RepSA on ImageNet-1K with 224$\times$224 image size. GPU denotes the throughput in image/s and NPU denotes the latency in iPhone 15 Pro.}
\begin{tabular}{ m{1.2cm}|>{\centering}m{1.4cm}|>{\centering}m{0.5cm}|>{\centering}m{0.8cm}|>{\centering}m{0.6cm}|>{\centering}m{0.6cm}|c } \hline
\multirow{2}{*}{Ablation} & \multirow{2}{*}{Variant} & Par & FLOPs & GPU   & NPU & \multirow{2}{*}{Top-1}  \\ 
                         &                           & (M) &  (G)  & img/s & ms  &        \\ \hline
\multirow{2}{*}{RepMSDW}& Multi-Scale & 9.3 & 1.12 & 3632 & 0.77 & 79.12 \\ \cline{2-7}
                        & Square      & 9.3 & 1.12 & 3632 & 0.77 & 79.0 \\ \cline{2-7}
                        & w/o Rep     & 9.3 & 1.12 & 3632 & 0.77 & 78.9\\ \hline
\multirow{3}{*}{RepSA} & RepMSDW$+$ MHSA & 12.7 & 1.31 & 3255 & 0.99 & 79.2  \\ \cline{2-7}
                       & RepMSDW$+$ SA  & 10.2 & 1.17 & 3491 & 0.80 & 79.6  \\ 
\hline

\end{tabular}
\label{tab:ablation_token_mixer}
\end{table}
\subsubsection{Combination in RAViT Backbone architecture}
Following the ablation of RepMSDW and RepSA, we also conducted an ablation study on the macro architecture. The investigation commenced with a comparison between the 3-stage (V1) and the 4-stage (V2) architectures. The findings from this analysis are detailed in Section \ref{subsec:abl-macro}, where V2 was adopted as the baseline configuration. Subsequently, we experimented with increasing the RepMSDW kernel size in stages three and four from $K=3$ to $K=7$ (V3), resulting in a 0.5\% enhancement in Top-1 accuracy, accompanied by a reduction in inference speed of 1\%, 2.5\%, and 10\% for NPU, EDGE, and GPU respectively. Incorporating RepSA in stage four (V4) yields a 0.9\% improvement in accuracy, but inference speed decreases by 3.8\%, 19\%, and 22\% on NPU, EDGE, and GPU, respectively. In the final ablation, we applied RepSA in both stages three and four, achieving a 1.7\% boost in accuracy, though inference speed slowed by 22.1\%, 44.9\%, and 78.6\% on NPU, GPU, and EDGE respectively. Ultimately, V4 was selected as the primary configuration for the RAViT backbone to balance between speed and accuracy.
\begin{table}[!ht]
\centering
\caption{Ablation of RAViT on ImageNet-1K with 224$\times$224 image size. M and A denote RepMSDW and  RepSA then followed by kernel size and number of blocks. C denotes a convolution followed by kernel size, stride, and number of blocks.}
\begin{tabular}{c|c|>{\centering}m{1.0cm}|>{\centering}m{0.9cm}|>{\centering}m{0.9cm}|>{\centering}m{0.9cm}|c} \hline
\multicolumn{2}{c|}{Config} & V1      & V2      & V3      & V4 & V5 \\ \hline
\multicolumn{2}{c|}{Stem} & C3S2$\times4$      & \multicolumn{4}{c}{C3S2$\times2$}\\ \hline
\multirow{4}{*}{Stage}  & 1  & M3$\times4$ & M3$\times2$ & M3$\times2$ & M3$\times2$ & M3 $\times2$  \\
                        & 2  & M3$\times16$ & M3$\times4$ & M3$\times4$ & M3$\times4$ & M3 $\times4$  \\
                        & 3  & M3$\times4$ & M3$\times12$ & M7$\times12$ & M7$\times12$& A7$\times12$ \\
                        & 4  & - & M3$\times4$ & M7 $\times 4$ & A7 $\times4$  & A7 $\times4$  \\ \hline
\multicolumn{2}{l|}{Param (M)}     & 10.0  & 9.18    & 9.3     & 10.2    & 10.9 \\ \hline
\multicolumn{2}{l|}{FLOPs (G)}     & 0.32  & 1.10    & 1.12    & 1.17    & 1.33  \\ \hline
\multicolumn{2}{l|}{GPU (img/s)}   & 13484 & 4059   & 3632    & 3316    & 2801 \\ \hline
\multicolumn{2}{l|}{NPU (ms)}      & 0.73  & 0.77    & 0.79    & 0.80    & 0.94 \\ \hline
\multicolumn{2}{l|}{EDGE (ms)}     & 10.0  & 9.8     & 9.9     & 11.7    & 17.5    \\ \hline
\multicolumn{2}{l|}{Top-1}         & 75.8  & 78.7    & 79.1    & 79.6    & 80.4 \\ \hline

\end{tabular}
\label{tab:ablation_macro}
\end{table}
\subsubsection{Fast-COS ablation}
Table \ref{tab:driver-scene} presents the ablation study of Fast-COS utilizing the RAViT variant as the backbone and RepFPN as the intermediary component of the original FCOS head detector. When compared to the standard FCOS employing ResNet-101 on the BDD100K dataset, incorporating RAViT-M26 enhances the detection accuracy for small $(AP_s)$ and large $(AP_l)$ objects by 4.8\% and 1.9\%, respectively. The system employing a reparameterizable RAViT backbone, particularly through the RepMSDW residual connection reparameterization, boosts the GPU inference speed by 62.1\%. By integrating RepMSDW into the FPN, notable improvements in $(AP_l)$, $(AP_m)$, and $(AP_s)$ are recorded at 2.7\%, 3.9\%, and 14.2\%, respectively. Moreover, the RepFPN version, functioning with merely three levels of feature extraction, enhances the GPU inference speed by 75.9\%. 

\subsection{Visualization Results}
Fig. \ref{fig:qualitativeobj} illustrates the comparison of detection results achieved by Fast-COS and the baseline algorithms. The images originate from the BDD validation set, which encompasses urban and suburban traffic scenes as well as traffic scenes in adverse weather conditions. Observing the detection results, it is evident that Fast-COS, through reparameterized multi-scale kernel size convolutions and reparameterized self-attention mechanisms, effectively resolves occlusion issues (such as occlusion between vehicles) and enhances detection performance relative to the baseline. This is particularly noticeable when detecting small objects like a signal light, traffic sign, or distant pedestrian. 

Through an expanded explanation, it is demonstrated within the first and second group of examples that, in urban traffic scenarios during daylight conditions, Fast-COS exhibits the ability to identify several traffic signs located at a considerable distance, in addition to an obstructed bus and automobile positioned in the background. In contrast, the baseline method is inadequate in its predictive capabilities. Moreover, in the third and fourth groups, Fast-COS effectively discerns a diminutive and partially hidden vehicle situated between two larger entities, even under nighttime conditions.

\section{Conclusion}
This paper presents Fast-COS, an innovative single-stage object detection framework optimized for real-time driving scene applications. By incorporating the novel Reparameterized Attention Vision Transformer (RAViT) as a hybrid transformer backbone along with the Reparameterized Feature Pyramid Network (RepFPN) for extracting features across multiple scales, Fast-COS achieves outstanding accuracy and computational efficiency. 

Key findings of this research indicate that the proposed framework markedly enhances the balance between accuracy and inference speed. With a Top-1 accuracy of 81.4\% on ImageNet-1K, RAViT outperforms other hybrid transformers such as FastViT, RepViT, and EfficientFormer in terms of GPUs, edge, and mobile inference speed. The combination of RAViT and RepFPN in constructing Fast-COS yields state-of-the-art performance on challenging driving scene datasets like BDD100K and TJU-DHD Traffic, outperforming conventional models such as FCOS and RetinaNet. The integration of RepMSDW and RepSA enhances both local and global spatial understanding while ensuring lightweight operations suitable for resource-limited hardware. 

Furthermore, extensive testing on GPUs and edge devices reveals the scalability and real-time efficiency of Fast-COS. The framework achieves up to a 75.9\% faster GPU inference speed and 1.38× increased throughput compared to leading models, making it an optimal choice for autonomous driving systems across various conditions and environments. Future research will focus on further architectural optimization for edge device hardware platforms, such as employing quantization for deployment.

\bibliography{egbib}
\bibliographystyle{ieeetr}

\ifCLASSOPTIONcaptionsoff
  \newpage
\fi

\end{document}